\documentclass[fleqn,10pt]{wlscirep}
\usepackage[utf8]{inputenc}
\usepackage[T1]{fontenc}
\usepackage{tikz,tcolorbox}
\usetikzlibrary{positioning, arrows.meta}
\title{Beware of ``Explanations'' of AI} 

  \author[1]{David Martens}
  \author[2,$\dagger$]{Galit Shmueli}
  \author[3]{Theodoros Evgeniou}
  \author[4]{Kevin Bauer}
  \author[5]{Christian Janiesch}
  \author[6]{Stefan Feuerriegel}
  \author[7]{Sebastian Gabel}
  \author[1]{Sofie Goethals}
  \author[8]{Travis Greene}
  \author[9]{Nadja Klein}
  \author[10]{Mathias Kraus}
  \author[11]{Niklas K\"uhl}
  \author[12]{Claudia Perlich}
  \author[13]{Wouter Verbeke}
  \author[14]{Alona Zharova}
  \author[15]{Patrick Zschech}
  \author[12]{Foster Provost} 
 
  \affil[1]{University of Antwerp, Department of Engineering Management, Antwerp, 2000, Belgium}
  \affil[2]{National Tsing Hua University, Institute of Service Science, Hsinchu, 30013, Taiwan}
 \affil[3]{INSEAD, Technology and Business, Fontainebleau, 77300, France}
 \affil[4]{Goethe University Frankfurt, Department of Information Systems, Frankfurt, 60629, Germany}
 \affil[5]{TU Dortmund University, Department of Computer Science, Dortmund, 44227, Germany}
 \affil[6]{LMU Munich and Munich Center for Machine Learning, Munich, 80539, Germany}
 \affil[7]{Erasmus University, Rotterdam School of Management, Rotterdam, 3062, Netherlands}
 \affil[8]{Copenhagen Business School, Department of Digitalization, Copenhagen, 2000, Denmark}
 \affil[9]{Karlsruhe Institute of Technology, Scientific Computing Center, Karlsruhe, 76131, Germany}
 \affil[10]{University of Regensburg, Faculty of Informatics and Data Science, Regensburg, 93053, Germany}
 \affil[11]{University of Bayreuth, Faculty of Law, Business and Economics, Bayreuth, 95440, Germany}
 \affil[12]{New York University, Department of Technology, Operations and Statistics, New York, NY 10012, USA}
 \affil[13]{KU Leuven, Faculty of Economics and Business, Leuven, 3000, Belgium}
 \affil[14]{Humboldt-Universität zu Berlin, School of Business and Economics, Berlin, 10099, Germany}
 \affil[15]{Leipzig University, Faculty of Economics and Management Science, Leipzig, 04109, Germany}

\affil[$\dagger$]{corresponding author: Galit Shmueli (galit.shmueli@iss.nthu.edu.tw)}

\begin{abstract}
Understanding the decisions made and actions taken by increasingly complex AI system remains a key challenge. This has led to an expanding field of research in explainable artificial intelligence (XAI), highlighting the potential of explanations to enhance trust, support adoption, and meet regulatory standards. However, the question of what constitutes a ``good'' explanation is dependent on the goals, stakeholders, and context. At a high level, psychological insights such as the concept of mental model alignment can offer guidance, but success in practice is challenging due to social and technical factors. As a result of this ill-defined nature of the problem, explanations can be of poor quality (e.g. unfaithful, irrelevant, or incoherent), potentially leading to substantial risks. Instead of fostering trust and safety, poorly designed explanations can actually cause harm, including wrong decisions, privacy violations, manipulation, and even reduced AI adoption. Therefore, we caution stakeholders to beware of explanations of AI: while they can be vital, they are not automatically a remedy for transparency or responsible AI adoption, and their misuse or limitations can exacerbate harm. Attention to these caveats can help guide future research to improve the quality and impact of AI explanations.
\end{abstract}

\begin{document}

\flushbottom
\maketitle
\thispagestyle{empty}

\section{Introduction}
As Artificial Intelligence (AI) systems\footnote{AI systems incorporate AI models, which are usually the product of machine learning. AI systems go beyond the mere prediction model, as they may use multiple models predicting different things, can have complex decision logic connecting the predictions to actions, and usually include  further business rules, including guard rails, black lists and white lists, etc.} are increasingly being used in organizations and society, and have a growing impact on decision making and our daily lives, trusting their behavior is crucial. 
State-of-the-art AI models, such as XGBoost or Random Forests for structured data or deep neural networks for unstructured data like images and text, can achieve remarkable predictive performance. However, their complexity renders them \textit{black boxes}, inherently uninterpretable by humans.\cite{martens2014,DBLP:journals/electronicmarkets/JanieschZH21,rudin2019} This opacity introduces challenges in areas such as error understanding and prevention, fostering user trust, and meeting regulatory requirements not only in form but also in spirit.\cite{NewEUAIreg} Unsurprisingly, the growing demand for safe and responsible AI adoption by organizations and individuals has heightened the need for clear, truthful and actionable explanations.

When AI outputs can be reliably traced back through the underlying processes, stakeholders---including regulators, managers, AI developers, and those affected by these outputs---can examine not only the outputs but also the inputs, data and methodologies driving them.  \cite{DBLP:journals/csur/DwivediDNSRPQWS23} 
Safeguards are indispensable as AI systems increasingly impact critical aspects of society, from healthcare and criminal justice to hiring and financial services, where the stakes can be high.\cite{NewEUAIreg}
However, there is often a notable mismatch between how and when explainable AI (XAI) methods are deployed and the ways humans comprehend and make decisions.\cite{poursabzi2021manipulating, alufaisan2021does} This gap arises largely from socio-technical complexities of interpreting AI outputs. \cite{DBLP:journals/ijinfoman/HermHWJ23,Babicetal2021} Current XAI research focuses predominantly on technical aspects, often overlooking the psychological, social, and contextual factors that shape human understanding and decision making. In fact, XAI research is arguably driven by the interests of the researchers rather than actual AI application needs, a phenomenon aptly described as the ``inmates running the asylum''. \cite{miller2017explainable} 

We advocate for a shift in both research and practice toward addressing critical questions necessary for enabling explanations to support the safe and responsible adoption of AI. We highlight how current explanations of AI decisions may create an illusion of accountability rather than fostering improved understanding, therefore also leaving the true reasons for the decisions inadequately scrutinized.
Accordingly, we caution ``Beware of Explanations of AI'': they may not (always) provide the remedy sought, and may even cause harm. To address these limitations, we explore potential root causes of the potential harms by identifying critical areas where increased awareness and scrutiny are needed to ensure that explanations contribute meaningfully to transparency, trust, accountability, and eventually the safe adoption of AI.

\section{A Socio-Technical Understanding of AI Explanations}
\subsection{What research in AI explanations provide us so far}
\textit{Explainable AI} can be defined as the research field that develops and applies algorithms to
explain AI system outputs.~\cite{martens2022data} 
Although the term \textit{XAI} is relatively recent (generally attributed to a DARPA project\cite{gunning2021}), the underlying need for explainability for AI system outputs has been recognized and addressed since AI systems started to make decisions that could affect people and the world.  The 1970s early medical AI system MYCIN %, in the 1970s, 
included a significant explanation component.\cite{MYCIN}  In the 1980s, Michalski\cite{DBLP:journals/ai/Michalski83} proposed the ``comprehensibility postulate,'' emphasizing that the results of machine learning should be interpretable in natural language.  In the 1990s, groundbreaking research was underway in understanding black-box machine learning models like neural networks.\cite{craven1995} Explainability is not merely a modern concern but a foundational aspect of building effective AI systems.

\emph{AI explanations} of system behaviors---meaning the AI system's predictions, decisions, or actions---aim to render the behaviors of complex model-driven AI systems comprehensible to humans. Put differently: AI explanations show why or how a system produced a specific output from a specific input. The design of AI explanations typically aims to address challenges associated with the black-box nature of AI, such as distrust, lack of accountability, insufficient error safeguarding, and limited knowledge discovery, without sacrificing the quality of the AI predictions, decisions, or actions. \cite{bauer2021explain,martens2022data}

The irony is that contemporary XAI methods aim to address the issue of complex black-box models by adding another technical layer of complex XAI methods.  Many XAI methods rely on a secondary algorithm designed to reveal (post hoc) important aspects of the behavior of an opaque primary AI system or the model driving it, such as a deep neural network. Thus, explainability is explicitly decoupled from the model itself, resulting in two distinct systems: the AI system (incorporating a model such as a neural network) and the algorithm responsible for generating explanations about the model or system's behavior (e.g., feature importance explanations like those produced by SHAP\cite{lundberg2017unified} or counterfactual explanation methods\cite{martens2014,wachter2018counterfactual}). These post-hoc explanation methods often are model-agnostic, meaning they can be applied to any predictive model regardless of its underlying architecture. Current explainability techniques are broadly categorized into \textit{global} methods, which aim to explain the overall behavior of the model, and \textit{local} methods, which focus on explaining specific predictions.\cite{martens2014} Popular approaches, such as counterfactual explanations, \cite{martens2014} saliency maps,\cite{martens2022data} and feature importance explanations,\cite{lundberg2017unified,ribeiro2016should} establish relationships between the input features $X$ and the model or system outputs $\hat{Y}$ to provide insights into why a particular output was---or was not---produced for a given input. By doing so, these methods enable individuals to better understand the model's input-output transformation.

However, despite decades of progress, much of current XAI research remains focused on developing technical solutions, often at the expense of answering, or even considering, the fundamental question: what constitutes a ``good'' explanation? For example, much of current XAI research remains centered on defining importance of a feature in some manner (e.g., highest gradient in target prediction,\cite{Selvaraju2017} largest approximate Shapley value,\cite{lundberg2017unified} largest coefficient in a linear surrogate model\cite{ribeiro2016should}) and then developing algorithms to generate those. This technical focus frequently neglects actual stakeholder goals, fails to adequately mitigate the challenges posed by black-box models, and leaves critical aspects of explanation quality, such as clarity, context, and relevance to stakeholders, insufficiently addressed or explored. This in turn limits the practical utility of XAI to address the important challenges we summarized at the outset. But why is it so difficult to simply define properly what a good AI explanation is, and develop an algorithm that universally solves this?

\clearpage
\subsection{What are (good) explanations?}
Definitions of what constitutes an explanation vary significantly across fields.\cite{garfinkel1982forms} Some scholars regard explanations as answers to ``why'' or ``how'' questions.\cite{wellman2011} Others conceptualize explanations as hypotheses linking causes to effects.\cite{einhorn1986judging,thagard2000explaining} 
Still others emphasize that explanations should convey a sense of causality that helps to make sense of a given phenomenon.

This diversity in definitions underscores the complexity of explanations. The role and purpose of explanations can become more apparent when viewed at a higher level through the lens of cognitive psychology. 
Central to the conceptualization of explanations is the concept of a \emph{mental model}, rooted in cognitive psychology. Mental models are ``\textit{all forms of mental representation, general or specific, from any domain, causal, intentional, or spatial},''\cite[p.193]{Brewer1987} encoding beliefs, facts, and knowledge. Explanations can thus be effectively understood as (mental) models that replicate the structure and dynamics of real-world phenomena, enabling both \textit{understanding} and \textit{prediction} of the phenomena in question.\cite{craik1967nature} Consistent with this view, Johnson-Laird emphasized that \textit{``understanding consists in having a working model of the phenomenon in your mind.''} \cite{johnson1983mental}

Now, what are good \textit{AI explanations}?
Wachter et al.~(2018) define an AI explanation as an attempt to convey the internal state or logic of an algorithm that leads to a decision.\cite{wachter2018counterfactual} This definition of an explanation as an `attempt' already indicates the complex nature of the problem. If we build on the prior, more general definitions, we can define a good AI explanation as one where there is a good fit between the mental model of the explanation recipient and the provided explanation.\cite{kayande2009how,martens2014explaining} This fit naturally depends on the particular recipient, the goal of the recipient in using the (X)AI system, and the particular circumstances,\cite{wanner23} which motivates why coming up with a ``good'' explanation is currently an ill-defined problem. This challenge is compounded by the technical fact that even given a particular notion of what makes for a satisfactory explanation, there can be multiple satisfactory explanations for the same input-output pair,\cite{martens2014explaining} and in addition, the set of explanations produced by some methods may not be robust to minor input changes. Let's dig into how a good explanation depends on the goal, stakeholder, and context.

\subsubsection*{Societal dimensions}
There are at least five key \textbf{purposes} for providing AI explanations\cite{martens2014}: (1) Increasing acceptance and likelihood of adoption of an AI system,\cite{benbasat} where explanations increase trust by validating the discovered patterns, and relatedly, improving the relationship between managers and system developers; (2) justifying individual system decisions (at inference/deployment time), e.g., to ensure or increase customer satisfaction and/or to satisfy ethical principles, such as providing user privacy and control;\cite{chen2017enhancing} (3) revealing erroneous reasoning (at inference time), and thereby allowing it to be corrected; (4) improving AI models and improving the machine learning that produces them, e.g., by adding guardrails to a model or by improving training data; (5)~improving long-term learning\cite{benbasat} by all stakeholders (learning about the world, the business domain, the company, the models’ reasoning processes, machine learning).

\begin{figure}
    \centering
    \includegraphics[width=0.74\linewidth]{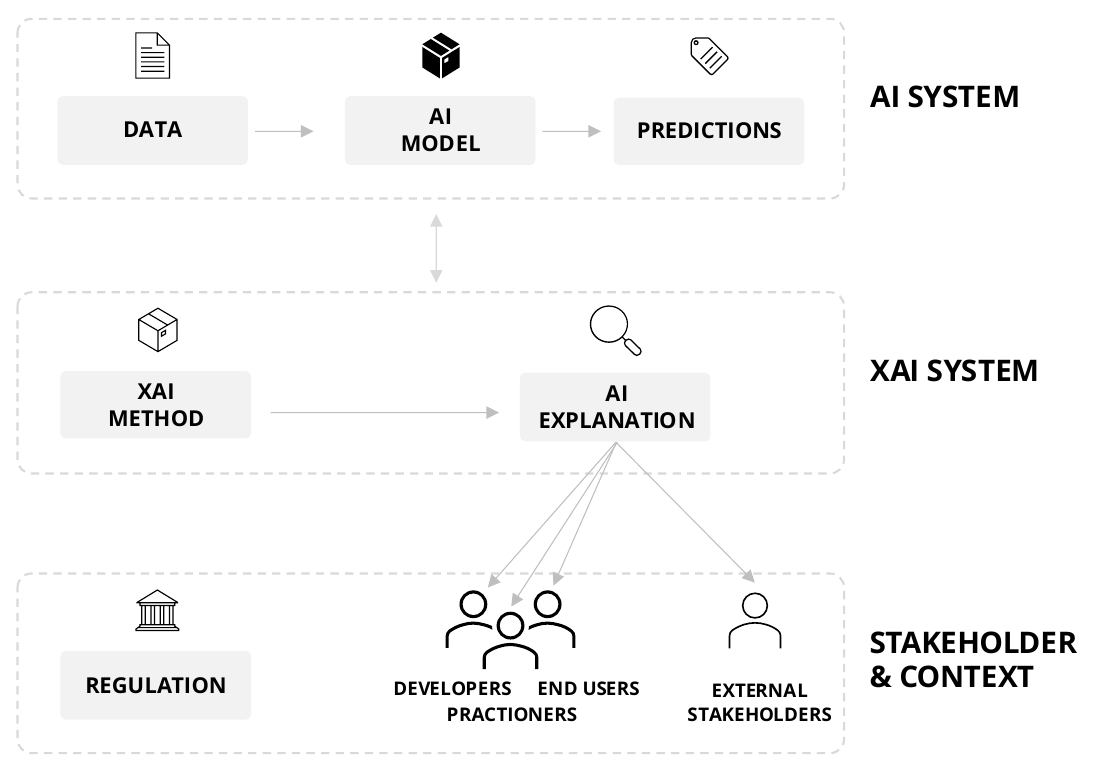}
    \caption{The XAI ecosystem includes the AI and XAI systems with their data and algorithmic components, as well as different human stakeholders and context.}
    \label{fig:xai-ecosystem}
\end{figure}

As explanations are never given or received in a social vacuum, it is pivotal to also consider the diversity in \textbf{stakeholders} throughout the entire machine-learning life cycle, encompassing an interconnected XAI ecosystem with partially competing interests (see Figure~\ref{fig:xai-ecosystem}). During the development phase, \textit{AI researchers}, \textit{data scientists}, and \textit{domain experts} rely on explanations to refine models, typically prioritize technical performance, and drive innovation. For these stakeholders, explanations serve as tools for debugging, feature selection, training data improvement, and understanding system/model behavior to improve decision making, overall reliability and model accuracy.

As AI systems move from development to deployment, the range of stakeholders expands notably. Practitioners such as (non-machine learning) \textit{engineers} and \textit{managers} use explanations to ensure operational reliability, troubleshoot unexpected system behaviors, and justify AI system outputs to \textit{customers, clients, users, decision-makers,} and \textit{other stakeholders}. In this phase, explanations are often required to bridge the gap between the technical intricacies of the AI system and the practical demands of organizational and business environments. Those affected by the decisions (which may include some of the aforementioned) represent another critical group to consider during deployment, with needs that vary based, for example, on their level of expertise and the context in which they interact with the system. For some, simplified narratives\cite{martens2025tell} may be sufficient to build acceptance of the system's actions, while others may require more detailed, technically rigorous justifications.

Beyond internal stakeholders and those directly affected by the system, there are other external stakeholders, including \textit{data subjects}, \textit{regulators}, and \textit{auditors}.  They introduce another layer of complexity. These stakeholders may demand explanations that demonstrate compliance with ethical, legal, and business standards, often requiring rigorous and transparent documentation. The needs of this group can diverge significantly from those of other stakeholders, emphasizing a level of formalization and accountability that may conflict with the simplicity sought by end-users, the trust-building needed by managers, or the efficiency and efficacy desired by developers.

And thirdly, the \textbf{context} is a key driver of what explanation is considered suitable or good from a societal perspective. For example, in high-risk applications, such as credit scoring or recruitment, much higher demands will be set to explanations than in low-risk applications as targeted advertising. This contextual issue is even the basis of the risk-based approach taken in the European AI Act.\cite{NewEUAIreg} These regulatory and societal requirements are yet another contextual dimension that will determine the needs for an AI explanation.

\subsubsection*{Technical dimensions}
A fundamental disconnect exists between the probabilistic nature of AI systems and the expectation that explanations provide crisp causal clarity. Further, even data scientists experienced in causal inference have trouble with the key explanatory distinction between causality at the AI system/model level and causality ``in the real world.'' The question, "why did the system deny me credit," is different from the question "why am I not credit-worthy."  AI systems do not necessarily faithfully represent underlying causal processes, and often do not need to (e.g., relying instead on accurate proxies).  Thus, while an AI model might produce accurate predictions using probabilistic methods, explanation recipients might be expecting a revelation of the latent causal mechanisms driving the actual phenomenon being modeled.  

This limitation is worsened by the selective nature of explainability methods. A single model (global explanation) or an individual prediction (local explanation) can yield multiple, distinct explanations.\cite{martens2014,barocas2020hidden,brughmans2024disagreement} While multiple explanations may be reasonable, such variability can make them seem inconsistent and unreliable, raising concerns about objectivity and interpretability. These methods, though capable of generating plausible narratives, may disappoint when users expect definitive or universal insights—an expectation often unrealistic (or even not desired\cite{thomas2022}) due to context-dependent explanation goals. Providing a single explanation is further complicated by competing priorities: granular explanations useful for developers may overwhelm end-users, while simplified versions may not meet regulatory needs. Achieving balance requires flexible, adaptable systems, presenting significant design and computational challenges.

\section{The XAI Causes and Caveats: Stakeholders Beware}
Given that the literature does not provide a clear understanding of what constitutes a universally ``good'' explanation, and given the wealth of different explanation goals and stakeholders, there are likely many context-dependent answers to that question. Thus, AI explanations in practice may not be tailored to the context. This leads to a set of root causes of ``bad'' explanations, some related to the quality of explanations in general, as noted in the literature (beyond AI), some due to particular characteristics of XAI, such as the ill-defined nature of the problem, the statistical nature of AI, or the key role of individual data in AI systems' decision making. % which may lead to data leakages. 
In particular, the following characteristics of AI explanations are root \emph{causes} of possible resulting harms, with an example from credit scoring to illustrate the problems. 
\clearpage
AI explanations may be:
\begin{enumerate}[label=\alph*.]
    \item \textbf{Unfaithful}: the explanation is inconsistent with the AI model and data, or inconsistent more generally. 
       \begin{tcolorbox}[colback=gray!5, colframe=gray!60, sharp corners] \underline{Example}:
    The explanation cites “employment stability” while this is not used by the model.
    \end{tcolorbox}
    \item \textbf{Irrelevant}: the explanation does not match the context or user needs, or provides excessive or insufficient detail.
       \begin{tcolorbox}[colback=gray!5, colframe=gray!60, sharp corners] \underline{Example}:
       The explanation cites all available features, while most do not have an impact on the decision made for the user, or even for the prediction score.
       \end{tcolorbox}
    \item \textbf{Unstable}: explanations produced by XAI may differ even when the same data and AI model are used.
        \begin{tcolorbox}[colback=gray!5, colframe=gray!60, sharp corners] \underline{Example}:
    The approximate SHAP explanation, generated twice with different random seeds, provides two different sets of `important' features.
    \end{tcolorbox}
    \item \textbf{Incoherent}: the explanation is too complex or unclear for the user, or is in conflict with the user's mental model without being clear why, or contains inconsistent elements that undermine understanding.
            \begin{tcolorbox}[colback=gray!5, colframe=gray!60, sharp corners] \underline{Example}:
    A SHAP explanation provided to a non-technical lay user, as the reason for rejected credit. \end{tcolorbox}
    \item \textbf{Revealing}: the explanation exposes personal, sensitive, or proprietary information.
                \begin{tcolorbox}[colback=gray!5, colframe=gray!60, sharp corners] \underline{Example}:
                A counterfactual explanation: ``If your income were higher than \$3,000, you would have been approved for the loan.'' reveals the bank's proprietary decision threshold for income.
                \end{tcolorbox} 
    \item \textbf{Unnecessary}: the explanation is unnecessary or redundant for the user's needs.
    \begin{tcolorbox}[colback=gray!5, colframe=gray!60, sharp corners] \underline{Example}:
                A SHAP explanation that lists the top 100 important features to a rejected applicant, while a counterfactual explanation would reveal that increasing income by \$200 alone would lead to the AI system accepting the applicant.
                \end{tcolorbox}
    \item \textbf{Cherry-picked}: the explanation selectively highlights evidence, ignoring contradictory information or sensitive or proprietary information not intended for sharing with the public.
    \begin{tcolorbox}[colback=gray!5, colframe=gray!60, sharp corners] \underline{Example}:
                A counterfactual explanation: ``If your income were higher than \$3,000, your job tenure 10 years longer, and your amount of loan \$100.000 lower, you would have been approved for the loan'' while the explanation ``If your gender was male instead of female you would be approved'' is intentionally not shown.
                \end{tcolorbox}
\end{enumerate}

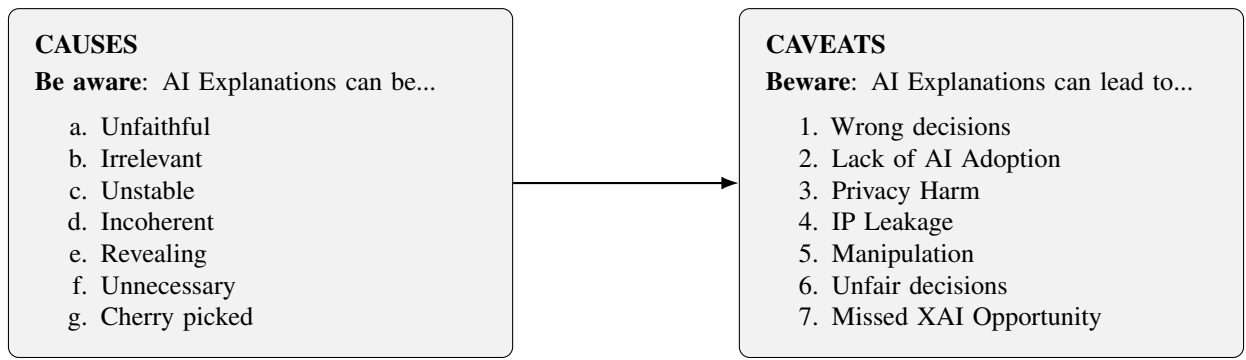
\begin{figure}
\begin{center}
    \begin{tikzpicture}
        % Define nodes
        \node[draw, fill=gray!10, rounded corners, inner sep=10pt, text width=6cm, align=left] (causes) at (0,0) {
            \textbf{CAUSES} \\[2pt]
            \textbf{Be aware}: AI Explanations can be... \\[5pt]
            \begin{enumerate}[label=\alph*.,nosep]
                \item Unfaithful
                \item Irrelevant
                \item Unstable
                \item Incoherent
                \item Revealing
                \item Unnecessary
                \item Cherry picked
            \end{enumerate}
        };
        
        \node[draw, fill=gray!10, rounded corners, inner sep=10pt, text width=6cm, align=left, right=3cm of causes] (outcomes) {
            \textbf{CAVEATS} \\[2pt]
            \textbf{Beware}: AI Explanations can lead to... \\[5pt]
            \begin{enumerate}[nosep]
                \item Wrong decisions
                \item Lack of AI Adoption
                \item Privacy Harm
                \item IP Leakage
                \item Manipulation
                \item Unfair decisions
                \item Missed XAI Opportunity
            \end{enumerate}
        };
        
        % Draw arrow between nodes
        \draw[thick, -Latex] (causes.east) -- (outcomes.west);
    \end{tikzpicture}
\end{center}
\caption{Due to the ill-defined nature of AI explanations, they can be of poor quality, leading to caveats to beware of.}
    \label{fig:bewarements}
\end{figure}

These limitations and inferior AI explanations can lead to actual harms---which we collectively call \emph{caveats}\footnote{Similar to the term \emph{caveat emptor}, meaning `buyer beware,' the implication is that the burden rests on the buyer to assess the quality of goods.}. Each root cause highlights an underlying characteristic of an explanation, such as unfaithfulness or irrelevancy, that (possibly in combination) gives rise to the caveats (see Figure \ref{fig:bewarements}). We argue that prudent users of XAI should beware of these caveats and their root causes. We now elaborate why. 
\clearpage

\begin{enumerate}
    \item \textbf{Explanations can lead to wrong decisions:} Automation bias, combined with convincing but poorly chosen explanations, can prompt stakeholders to trust and follow flawed AI outputs, even when those may be harmful to them. This overtrust can lead to wrong decisions with serious consequences, such as a doctor relying on flawed AI advice for treatment, 
    simply because the explanation is not in disagreement with people's own mental model (while the AI model is). The explanation can thus unintentionally create an illusion of understanding, where users believe they grasp the underlying logic of a predictive model despite its complexity remaining beyond their comprehension.\cite{rozenblit2002misunderstood, cabitza2024explanations} This illusion can be further compounded by hindsight bias, where users, after receiving an AI explanation, falsely assume they could have predicted the output.\cite{keil2006explanation} Finally, explanations that fail to offer actionable insights may also result in confusion or fail to guide decision making, \cite{liao2022connecting} once more potentially leading to poor decisions. 
    \begin{tcolorbox}[colback=gray!5, colframe=gray!60, sharp corners] \underline{Example}:
    A judge using an AI system to predict probability of recidivism, throws out a plea agreement between the prosecutor and the defendant, purely based on the AI explanation indicating that the most important features in the prediction are the defendant's unstable employment, lack of community support, and his criminal record. All this while the AI model predicts a \underline{low} risk of recidivism.
    \end{tcolorbox}
    
    \item \textbf{Explanations can lead to lack of adoption of AI and missed opportunities:} Unfaithful or overly complex explanations can cause misplaced distrust, leading users to question quality AI outputs that may in fact be helpful. While explanations can build trust and positively impact AI adoption, poorly chosen explanations can have counterproductive effects. When AI explanations fail to align with stakeholders' mental models, they may hinder rather than help adoption, even if the AI system has the potential to improve outcomes. Explanations that are irrelevant or incomprehensible can exacerbate this distrust.\cite{papenmeier2019model} They may even provide users with an excuse to discard relevant or helpful AI outputs that they do not agree with for strategic or subjective reasons. 
\begin{tcolorbox}[colback=gray!5, colframe=gray!60, sharp corners] \underline{Example}:
A credit officer receives SHAP explanations of the 20 most important features, but doesn't understand the
meaning of the SHAP values, and is overwhelmed with the complexity of the set of features. In the officer's mind,
income of the applicant is the only reason for rejection. Therefore, the credit officer does not trust the AI model to
work well and ignores its predictions, though it would substantially improve its lending practices.
    \end{tcolorbox}
    
    \item \textbf{Explanations can lead to privacy harms:} AI explanations, especially when generated interactively or repeatedly for multiple input data points, could expose personal, sensitive information that was used to train the AI model. For instance, when explanations are generated using instance-based strategies, they may reveal sensitive details about individuals in the training data. A notable risk is the explanation linkage attack, where adversaries link counterfactual explanations to background information, potentially identifying and extracting private attributes of individuals.\cite{goethals2023privacy} Another type of attack is a membership inference attack, where it is determined whether a specific individual is part of the training data.\cite{shokri2020exploiting, naretto2022evaluating} 
    \begin{tcolorbox}[colback=gray!5, colframe=gray!60, sharp corners] \underline{Example}:
 The hospital technician repeatedly requests explanations from the AI diagnostic system for the same patient. Each time, different features such as age, family history, or genetic markers are highlighted. By combining the explanations with publicly known patient attributes, the technician can re-identify the person as a certain colleague and can infer that the colleague carries a high-risk genetic condition.
   \end{tcolorbox}
   
    \item \textbf{Explanations can lead to IP leakage:} Not only sensitive information about persons might be revealed, but also proprietary information can be revealed through AI explanations, providing competitors with sensitive details that undermine the explainer's strategic position. For example, trade secrets embedded in the AI model or unique patterns in the data may be exposed via explanations, giving rivals an unintended advantage. Also, given enough explanations, adversaries may be able to reverse engineer the AI model used.\cite{aivodji2020model, yan2023explanation} This harm is rooted in AI explanations being overly transparent, where they reveal more than intended about the underlying system. 
\begin{tcolorbox}[colback=gray!5, colframe=gray!60, sharp corners] \underline{Example}:
A credit card company uses AI models to detect fraud based on transaction amount, frequency, location, and account history. Flagged transactions trigger detailed explanations, such as “round amounts” (e.g., exactly \$500.00) or “abnormal overseas transfers” (e.g., a transaction in China within 10 hours of an in-store purchase in the U.S.). These explanations help fraudsters adapt, by avoiding round amounts, splitting transfers, spacing out transactions, or using VPNs, allowing them to evade future detection while still committing fraud.
\end{tcolorbox}
    
    \item \textbf{Explanations can be used to manipulate stakeholders:} Explanations, particularly when misused by malicious actors or those with conflicting incentives, can become tools for manipulation. Those actors might intentionally mislead stakeholders, possibly at scale. Given the multiplicity of possible explanations, providers could exploit this flexibility to manipulate while still claiming correctness in a technical XAI sense.\cite{goethals2023manipulation} The possibility of cherry-picked or unfaithful AI explanations exacerbate this risk, as they allow malicious actors to control narratives for personal gain. 
\begin{tcolorbox}[colback=gray!5, colframe=gray!60, sharp corners] \underline{Example}:
After an accident involving its autonomous vehicle, the manufacturer releases explanations pointing to poor road conditions or human errors as the primary causes. The explanation omits any mention of the vehicle's delayed braking system, and as such shifts blame away from the AI system. The explanations can still be correct, but are not the only possible explanations or contributing factor for the accident.
\end{tcolorbox}
   
    \item \textbf{Explanations can lead to unfair decision making:} Explanations may support intentional or unintentional unethical behavior, such as discrimination. For example, AI explanations might cherry-pick details to intentionally mislead, such as omitting gender or a clearly correlated feature when the underlying model makes decisions that are unfavorable, for example, to a protected class. Similarly, they might fabricate unfaithful explanations via a surrogate model that are not aligned with the actual AI model's reasoning, such as falsely attributing a decision to insufficient income when income has no impact on the AI system's output. This kind of misuse can also include efforts to avoid liability (fairwashing\cite{aivodji2019fairwashing}), or provide minimal information to deter further inquiries (``keeping users at bay''). These tactics exploit the multiplicity of possible explanations, enabling providers to choose (or manipulate) narratives while maintaining an appearance of technical correctness and compliance with AI regulations. 
\begin{tcolorbox}[colback=gray!5, colframe=gray!60, sharp corners] \underline{Example}:
Automatic discarding of explanations that reveal the use of sensitive information that is illegal to use in credit scoring. For example, in many countries, it is illegal %not allowed 
to use information on gender in credit risk modeling. To conceal the use of this information, banks could cherry-pick explanations that do not include any sensitive attributes. These alternative explanations allow the bank to maintain a misguided perception of fairness and transparency.\end{tcolorbox}

    \item \textbf{Explanations can lead to missed opportunity for XAI value creation:} Overemphasis on explainability can lead to significant resource allocation for AI explanations that may not be needed in a given context. For example, providing detailed counterfactual explanations for low-stakes decisions, such as product recommendations, or to users who do not desire them (e.g. those who were provided credit), can detract from other priorities. This harm is linked to caveats such as irrelevant or not-needed explanations, where the provided explanations are misaligned with user needs or the decision's importance. Developing processes to assess when and how explanations add value is critical to avoid wasting resources.
    \begin{tcolorbox}[colback=gray!5, colframe=gray!60, sharp corners] \underline{Example}:
    A credit card company uses AI models to detect fraud based on transaction patterns. To enhance transparency, it provides SHAP-based explanations not only for flagged fraudulent transactions but also for all non-fraudulent transactions, informing customers why their transactions were not flagged. Given that each explanation costs \$0.01 in computing power and the company processes 100 million transactions daily, this results in excessive costs for explanations that customers do not want.  \end{tcolorbox}
\end{enumerate}

\section{Conclusion} 
Developers and users of AI and XAI should beware of the caveats of AI explanations and their root causes, keeping in mind the various ways explanations can actually cause harms. While XAI holds substantial promise for enhancing trust and transparency in an age of complex AI systems---and can do so if employed sensibly---it also introduces new or scales up existing socio-technical challenges. Research and development of XAI is still not focused on what constitutes a ``good'' explanation, which risks oversimplification or misuse. We acknowledge that XAI is vital, but we caution against indiscriminate adoption. Instead, we urge all stakeholders, be they consumers, data scientists, researchers, managers, or policy makers, to recognize the complexity and caveats. Approach XAI with caution, as AI explanations are not a silver bullet for transparent AI-driven decision making.

Moving forward, instead of merely focusing on technical XAI methods and tools, we should aim for explanations that genuinely serve their intended purpose as the harm from a poor explanation could exceed that of a black-box prediction. The caveats underscore the importance of interdisciplinary collaboration, combining insights from machine learning, psychology, philosophy, information systems, and human-computer interaction. The goals of combining these perspectives should be to better define the problem given the purpose, stakeholder, and context, to innovate methods that avoid or reduce the effects of the various caveats, to produce thoughtful regulation designed to ensure that what is desirable is indeed achieved with minimal negative side effects, and to heighten awareness to ensure explanations are applied meaningfully and responsibly. Currently, end-users must do their own due diligence to assess the quality of explanations given to them. However, our position is consistent with legal proposals \cite{yew2022penalty} that would shift the burden of proof on XAI developers to preemptively disclose the context-sensitive risks and harms of AI explanations to regulators.  \textit{Caveat Procurator!}

\section*{Acknowledgments}
This work was inspired by Dagstuhl Seminar 24342; we thank Schloss Dagstuhl for providing an inspiring research environment.  

\bibliography{explanations}

\end{document}